\documentclass{article}

\usepackage{arxiv}

\usepackage{hyperref}

\usepackage{inputenc,framed,color,amsmath, amssymb}
\usepackage{natbib}
\usepackage{algorithm}
\usepackage{algorithmic}
\usepackage{tcolorbox}
\usepackage{booktabs,subcaption} 
\newcommand{\esha}[1]{\textcolor{cyan}{\textbf{}}}
\newcommand{\melissa}[1]{\textcolor{blue}{\textbf{}}}
\newcommand{\marcello}[1]{\textcolor{green}{\textbf{}}}
\newcommand{\saeed}[1]{\textcolor{yellow}{\textbf{}}}
\newcommand{\aux}{\mathsf{aux}} 
\newcommand{\Dtrain}{\mathsf{D_\mathsf{target}}}
\newcommand{\Dshadow}{\mathsf{D_\mathsf{shadow}}}
\newcommand{\Din}{\mathsf{D_\mathsf{in}}}
\newcommand{\Dout}{\mathsf{D_\mathsf{out}}}
\newcommand{\myparagraph}[1]{\paragraph{\textbf{#1}}}

\author{{Saeed Mahloujifar} \\
	Princeton University\\
	\texttt{sfar@princeton.edu} \\
	\And
	{Huseyin A. Inan} \\
	Microsoft Research\\
	\texttt{huseyin.inan@microsoft.com} \\
	\And
	{Melissa Chase}\\
	Microsoft Research\\
	\texttt{melissac@microsoft.com} \\
	\And
	{Esha Ghosh} \\
	Microsoft Research\\
	\texttt{esha.ghosh@microsoft.com} \\
	\And
	{Marcello Hasegawa} \\
	Microsoft Corporation\\
	\texttt{marcellh@microsoft.com} \\
}

\date{}

\renewcommand{\headeright}{}
\renewcommand{\undertitle}{}

\begin{document}
\title{Membership Inference on Word Embedding and Beyond} 

\maketitle

\begin{abstract}

In the text processing context, most ML models are built on word embeddings. These embeddings are themselves trained on some datasets, potentially containing sensitive data. In some cases this training is done independently, in other cases, it occurs as part of training a larger, task-specific model. In either case, it is of interest to consider membership inference attacks based on the embedding layer as a way of understanding sensitive information leakage. But, somewhat surprisingly, membership inference attacks on word embeddings and their effect in other natural language processing (NLP) tasks that use these embeddings, have remained relatively unexplored. 
 
In this work, we show that word embeddings are vulnerable to black-box membership inference attacks under realistic assumptions. Furthermore, we show that this leakage persists through two other major NLP applications: classification and text-generation, even when the embedding layer is not exposed to the attacker. We show that our MI attack achieves high attack accuracy against a classifier model and an LSTM-based language model. Indeed, our attack is a cheaper membership inference attack on text-generative models, which does not require the knowledge of the target model or any expensive training of text-generative models as shadow models.
\end{abstract}

 \section{Introduction}
 There has been a rich body of work \citep{Mireshghallahsurvey20, Tanuwidjaja19, hu2021membership} that investigates Machine Learning (ML) pipelines through the lens of privacy and information leakage. This body of work largely investigates the question of what information ML models capture and expose beyond the task at hand. A natural attack metric that is commonly used to understand the extent of information leakage from a ML model is \emph{Membership Inference} (MI) attack. In MI attacks, an attacker is given black/white/grey box access to a ML model and aims to find out if a particular set of data was used in training the ML model. MI has been investigated in many domains such as vision \citep{shokri17, he2020segmentationsleak}, generative adversarial networks \citep{LOGAN, GAN-Leaks}, graph neural networks \citep{he2021nodelevel, olatunji2021membership} among many others.
 
 \myparagraph{Word Embedding} In the text processing context, most ML models are built on word embeddings, which provide a mapping from words in a dictionary to vectors in an embedding space; these vectors can then be used as the input for the rest of the model. These embeddings are themselves trained on some dataset, potentially containing sensitive data.  In some cases this training is done independently, in other cases it occurs as part of  training a larger, task specific model. In either case it makes sense to consider membership inference attacks on this embedding layer, where the adversary tries to determine if a dataset was used in training the embeddings.  
 
Membership inference attacks in this setting, i.e., in the context of embeddings and its effect in other natural language processing (NLP) tasks that use embeddings, are relatively unexplored\footnote{To the best of our knowledge~\cite{embeddingccs} is the only work that explored MI for word embeddings. We give a detailed comparison with~\cite{embeddingccs} in Section~\ref{sec:compare}}. Given the rich set of applications where word embeddings appear as the first block of deep neural models, this is somewhat surprising.


 
 In this work, we investigate MI attacks on word embeddings and its two major applications in the NLP domain (namely, classification and text-generation). We assume label-only (also known as black-box) access to the target model. We note that, sometimes grey-box attacks (where the adversary gets confidence scores) have been referred to as black-box attack in the literature. But here we mean black-box access in the sense that the adversary only sees the predicted label. Grey-box attack model is strictly less general than black-box attack model since score-based attacks cannot be applied when the model only exposes the predicted label, which is often the case through the API boundary in production model deployment. In this work, we ask the following questions:
 
 \begin{itemize}
     \item Given black-box access to a word embedding function can a semi-honest attacker\footnote{A \emph{semi-honest} attacker refers to a passive adversary who follows the protocol and does not inject any malicious data in the training phase. This term is commonly used in the cryptographic literature}. infer if her data was used in training the embedding?
     \item If the semi-honest attacker has label-only access a model built on an embedding (instead of the embedding layer itself) that uses some embedding function, can the attacker still infer if her data was used in training the model?
 \end{itemize}

We answer both these questions in positive. We choose two different NLP tasks: a spam classifier as a classification task and a language model as a text-generation task to show that the information leakage persists even when the embedding layer is not directly exposed. Perhaps the more surprising of these two applications is the spam classifier in which the attacker only gets access to the prediction label of spam or not spam.


Our attack tries to exploit a fundamental property of any ``good'' embedding function. A good embedding is expected to preserve some semantic meaning and relations of the underlying objects. In the case of word embeddings, this means, a good model is expected to capture semantic relationships between words. For example, the words \emph{king} and \emph{male} are expected to be closer together in the embedding space as compared to the pair \emph{king} and \emph{female}. Our attacks exploit this property in a clever way. We try to find some special word pairs in the attacker's dataset that are not semantically close generally. Note that we still want all these special words to belong to the vocabulary of the training dataset to rule out trivial attacks. Then we try to infer if those words appear significantly closer in the embedding space, which would be a strong signal that the attacker's data was used in the training.

To give an intuition about these special word pairs, imagine that the attacker is trying to infer if emails from the year late 2020 were used in training an embedding function. The words \emph{remote} and \emph{work} would appear much closer together in the emails from 2020 than from before. So (remote, work) is a special word pair, indicative of the email set of late 2020. 

We build on this intuition to first attack the word embedding function directly. Our attack is can achieve success rates of above $90\%$ in doing membership inference attack against word embeddings. Surprisingly, we show that this leakage from word embedding is transferable to the classification and text-generation NLP tasks.


Our attack provides a new path for attacking NLP models. For instance, our MI attack on text-generative language models does not require the knowledge of the target model or any expensive shadow model training (as opposed to previous work). While we show the success of our attack on an LSTM-based \citep{Sundermeyer_lstmneural} language model, a by-product of our attack is a cheap MI attack on language models, which we expect, would generalize to MI attacks on other large transformer-based models \citep{transformer} for which shadow model training would be highly costly. 
 
\myparagraph{Contributions}

\begin{enumerate}
    \item We provide a novel membership inference (MI) attack against word embeddings that exploits an important property of any good word embedding: that it captures semantic relationship of words and that words adjacent in training data are considered semantically close. We show the success of our attack against Word2Vec which is one of the most commonly used word embedding algorithms. Our attack can achieve MI accuracy around $90\%$. (See Section \ref{sec:embed_attack}).
   
   \noindent Note that, in contrast to previous work~\cite{embeddingccs}, our attack works in a very realistic setting where the attacker does not have knowledge of or sample access to the exact training data distribution.
   
   \item Then we show how to extend our attack to membership inference against a spam classifier and a text-generative model when the attacker only has black-box access to these models. We show that our MI attack achieves $65-90\%$ attack accuracy against the classifier model (Section~\ref{sec:classifier_attack}) and $70\%$ attack accuracy against LSTM-based language model  (Section~\ref{sec:tg_attack}). 
   
   \item A secondary contribution of our attack is providing a new approach for MI on NLP models. Our MI attack on text-generative language models for instance does not require the knowledge of the target model or any expensive shadow model training as opposed to previous work, while still providing successful attack performance. We further note that our attack transfers through LSTM-based language model where the trained embedding layer is not based on Word2Vec. Therefore, we expect our MI attack to generalize to other large transformer-based models for which shadow model training would be highly costly.
\end{enumerate}

\myparagraph{Organization} The paper is organized as follows. In Section~\ref{sec:threatmodel} we define the threat model and formal security experiments. We discuss our attack on word embedding in Section~\ref{sec:embed_attack}, on the classifier in Section~\ref{sec:classifier_attack} and on the text-generative model in Section~\ref{sec:tg_attack}. Finally, we discuss related work in Section~\ref{sec:related} and conclude in Section~\ref{sec:conclusion}.
 
 \section {Security Model} \label{sec:threatmodel}
  In this paper, we focus on black-box membership inference attacks, introduced in \cite{shokri17}, in which an attacker can query a model with the goal of determining whether a particular dataset was included in the training data.
 
 More formally, we consider a meta-distribution $\mathcal{D}$, from which we will sample $n$ distributions $\mathcal{D}_i$.  Intuitively, $\mathcal{D}$ captures the distribution over users  and $\mathcal{D}_i$ captures the distribution over datasets of one particular user.
 
 We consider an adversary that targets a particular user with dataset ${D^*}$.\footnote{This is somewhat different from the approach generally taken to define membership inference, but we think this is beneficial because it allows us to discuss the vulnerability of a particular user dataset, rather than just an average vulnerability over all users.} First, $n$ distributions are chosen from $\mathcal{D}$, one corresponding to each user who provides training data.  Specific datasets $D_1,\ldots, D_{n}$ are chosen from the respective distributions. Dataset ${D}^*$ is given to the adversary, and a model is trained on either datasets $D_1, \ldots D_{n-1}, D^*$, or $D_1,\ldots, D_n$.  The adversary makes a series of black-box queries and must determine which occurred. We formally model this as a security experiment in Figure~\ref{fig:userdsattack}.


\begin{figure}
\begin{framed}
$D^*, \mathcal{D}$, and $n$ are parameters of the game.
    \begin{enumerate}
         
        \item Select a bit $b\in \{0,1\}$ uniformly at random. 
        \item Sample $n$ distributions $\mathcal{D}_1,\ldots, \mathcal{D}_{n}$ from $\mathcal{D}$. 
        
        \item Sample $D_i \gets {{\mathcal{D}}}_i$ for $i = 1, \ldots n-1$
        \item If $b=0$, set $D_n = D^*$. If $b=1$, sample $D_n\gets \mathcal{D}_n$
        \item $D^*$ is given to the attacker, $A$. 
        \item $A$ is assumed to have some other parameters which we will denote as $\aux$
        \item Train a model $M\gets L(D_1, D_2,
        \ldots, D_n) $.
        \item $A$ adaptively queries the model on a sequence of points $x_1,\dots,x_t$ and receives $y_1 =M(x_1),\dots, y_t=M(x_t)$.
        \item $A$ then outputs a bit $b'$ and wins the game if $b=b'$.
    \end{enumerate}
\end{framed}
\caption{Security Experiment for attack against distribution of users}
\label{fig:userdsattack}
\end{figure}

\paragraph{Attack success against user dataset $D^*$.}  To evaluate how vulnerable a particular user dataset is to membership inference attack, we run the experiment in Figure~\ref{fig:userdsattack} many times and compute the success probability. 

\paragraph{Attack success against distribution $\mathcal{D}$ of users.}  We can also compute an average success metric for the attack against a meta distribution $\mathcal{D}$ by sampling many user distributions, and then sampling a specific dataset for each, computing the estimated vulnerability of each dataset, and then averaging the results\footnote{If we use only one repetition to estimate the success against each user, and if all $D_i$'s represent the same distribution, this is equivalent to the definition in \cite{DBLP:conf/csfw/YeomGFJ18}.}.

Computing this attack metric can be quite expensive for more complex target models, in that it requires training the target model several times for each user. An alternate way to compute this latter success metric is as follows\footnote{A similar approach is taken in \cite{song19}.}: 

We begin with a dataset which contains many users (say $n$). We train a model on a dataset containing the data from a random $n/2$ subset of the users. Then we run the attack for each user in the dataset, giving the attacker the goal of determining whether that user was in the chosen subset.  We repeat this process several times to minimize the effect of randomness, and then report the attack's average success rate.
We formally define the security experiment in Figure~\ref{fig:distattack}.
\begin{figure}
\begin{framed}
$\mathcal{D}$ and $n$ are parameters of the game.

    \begin{enumerate}
        \item Pick $n$ distributions $\mathcal{D}_i$ for $i = 1, \ldots n$ where $n$ is even.
        \item $D_i \gets \mathcal{D}_i$
        \item Run the following several times:
        \begin{enumerate}
            \item Initialize a bit vector $\textbf{b}= b_1, \ldots, b_n = 0^n$
            \item Pick a random $n/2$ subset of these datasets. Let us denote these datasets $D'_1, D'_2,\ldots, D'_{n/2}$.
            \item Set the corresponding bits in $\textbf{b}$ to 1. 
            \item Train a model $M\gets L(D'_1, D'_2,\ldots, D'_{n/2}) $.
            \item For $i\in [n]$:  Run the attack as follows:
            \begin{itemize}
                \item The attacker is given $D_i$ for $i \in [n]$.
                \item The attacker is also given some other parameters which we will denote as $\aux$
                \item The attacker adaptively queries the model on a sequence of points $x_1,\dots,x_m$ and receives $y_1 =M(x_1),\dots, y_m=M(x_m)$.
                \item The attacker outputs a bit $b'_i$ indicating whether it thinks $D_i$ was included in the training set.
            \end{itemize}
            \item  Compute S=  $\sum_{i=1}^{n} (b'_i = b_i)/n$
        \end{enumerate}
        \item We compute a metric for the success of the attack by averaging the values of $S$ obtained in each of the runs.  This roughly captures the average success probability of the attacker attacking a randomly selected user.
    \end{enumerate}
\end{framed}
\caption{Security Experiment for attack against a user dataset}
\end{figure}\label{fig:distattack}


\myparagraph{Variations} The security experiments are generic and allows for implementation by varying the parameters of the experiment. We discuss this below.
\begin{itemize}
\item Access to the distributions: $\aux$ could include sample access to the distributions $\mathcal{D}_1, \ldots \mathcal{D}_n$ or include sample access to distributions similar to $\mathcal{D}_1, \ldots \mathcal{D}_n$, but not those. For example, $\mathcal{D}_i$ could be users sampled from the Enron email distribution while a similar distribution could be the Avocado email distribution.
\item Target model types: Model $M$ could be three different types in our attacks (embedding, discriminators, and tex-generative models), but the security experiment can allow any model. The queries and outputs in the penultimate state of the security game is decided by the type of model. For example, if $M$ is a word embedding, the input is a word and the output is an embedding vector.  If $M$ is a text generative model, the input is a text sequence and the output is the predicted next word.
\item Training on a subset of the data: As a variation, we can consider the case where the training algorithm only uses a random subset of each user/tenant's data.  This makes the attacker's task somewhat more difficult because, while the attacker knows the dataset of the user/tenant in question, it does not know which subset of that user/tenant's data will be used.  This more accurately reflects what happens in many real world training contexts.
\end{itemize}

  \section{Word Embedding Attack}
 \label{sec:embed_attack}
Word embedding are maps where each word $w$ from a dictionary $Q$ is represented as a vector of fixed dimensions, $d$. We denote the vector as $v_w \in \mathbb{R}^d$. Word embedding vectors capture the semantic meaning of words using a distance metric. Words with similar semantic meanings will have small cosine distance in the embedding space.
 As formally defined in previous section, the goal of the attacker is to find out whether a set of examples corresponding to a user has been used to train a target word embedding. To reach this goal, the adversary must find a signal in the input-output behavior of the embedding that reveals this information. A general framework to find such signals is shadow model training procedure where the adversary trains a batch of embeddings $m_1,\dots,m_k$ as shadow models, with the data from the target user and $m'_1,\dots,m'_k$ without the data from the target user and then tries to find some statistical disparity between these models. 
 
When we are in the black-box setting, the statistical disparity should be manifested in the black-box behavior of the model which makes the job of adversary even harder. One naive way to find these statistical differences is to query the trained model on a large set and use the responses as representation for the model. Then, machine learning can help to identify if there is any difference between the responses that came from the set of models that had the target user data and models that did not have the target user's data \citep{shokri17}. Indeed, this is the first idea that comes to mind when trying to attack word embedding, just query the model an all words in the dictionary to get a representation of model in  $R^{Q\cdot d}$ (where $Q$ is the number of words in the dictionary and $d$ is the dimension of the embedding space) and then figure out the membership of the target user by investigating this high dimensional representation. There are two issues with this approach that we explain below. This approach is expected to be effective as we expect that the membership of the target data to have its largest effect on the representation of the words that are present in the target dataset.

\myparagraph{High Dimensions} One major problem with the approach proposed above is that the dimension of the model representation is very large for machine learning models to handle. The size of dictionary is huge and that will be multiplied by the dimension of the embedding model. One way to fix this is instead of working with all the words in the dictionary, we can sub-sample a small fraction of words and use the responses from the model on them as the representation of the model. This is a naive way of reducing the dimension of the representation that is independent of the target user's data. A smarter way to the sub-sampling is to sample the words based on the target user's data. For example, one can query only the words that exist in the target dataset and this reduces the dimension significantly. 
\melissa{And we expect this to be effective because...}

\myparagraph{Randomness in the embedding space} The goal of word embedding is to map the words into a geometric space in a way that preserves the semantic closeness. For example, one would expect the representation of the word ``dog'' to be closer to representation of ``cat'' than that of ``cake''. This is indeed one only requirement that one will expect from an ideal word embedding. Now, for an embedding model $m$ if one considers a transformation of $\pi$ that maps $m$ to $m'=\pi(m)$ such that the relative distances between pairs of words are preserved then $m'$ will be as good as $m$ in representing words. There are many possible transformation that preserve distance such as rotation or shifting. In that sense there is a high entropy in the representation of words which makes the job of adversary harder in identifying a statistical pattern. 

\myparagraph{Our Attack} In order to handle the two issue mentioned above we propose an attack that uses $l_2$ distance between the representation of words in the users email data. This way we resolve the second issue as we look at the relative distance between words and we expect the randomness of the word embedding not to change the relative distances between pairs of words. The reason behind this expectation is the fact that word embeddings are designed to create correlation between the semantic distances of words and their Euclidean distance in the embedding space. This technique also resolves the first issue mentioned above. We only look at the pairs of words in the target dataset which reduces the dimension from $Q\cdot d$ in the naive representation to $\binom{T}{2}$ where $T$ is the number of words in the target dataset. In order to further reduce the dimension, in our attack, we actually just consider the consecutive pairs of words which reduces the dimension to $T-1$. The reason behind this is that we expect the presence of the target data in the training set should have a more significant effect on pairs of words that are closer to each other. Below, we describe our attack algorithms.
\subsection{Attack as per the Security Experiments}\label{ssec:embedding_expt}
\begin{framed}
$D^*$ is one user's data from Enron email data.
    \begin{enumerate}
         
        \item Select a bit $b\in \{0,1\}$ uniformly at random. 
        \item Sample $n$ distributions $\mathcal{D}_1,\ldots, \mathcal{D}_{n}$ from $\mathcal{D}$. In our experiments all these distributions are Enron email distribution.
        \item Sample $D_i \gets {{\mathcal{D}}}_i$ for $i = 1, \ldots n-1$. Here each $D_i$ represents a user's data. Since Enron data does not have a notion of user, we emulate a user by selecting random emails from the dataset. \label{line:whichdist}
        \item If $b=0$, set $D_n = D^*$. If $b=1$, sample $D_n\gets \mathcal{D}_n$
        \item $D^*$ is given to the attacker, $A$. 
        \item $A$ is assumed to have a sample access to a shadow distribution $\aux=\mathcal{D}_{\mathsf{shadow}}$. We describe what shadow distributions we use in Section~\ref{ssec:embedding_attack}.
        \item Train a Word2vec embedding $M\gets L(D_1, D_2,
        \ldots, D_n) $.\label{line:whichmodel}
        \item $A$ adaptively queries the model on a sequence of points $x_1,\dots,x_t$ and receives $y_1 =M(x_1),\dots, y_t=M(x_t)$. We discuss how $A$ prepares these query points in Algorithm~\ref{alg:embedding_prep}\label{line:prep}.
        \item $A$ then outputs a bit $b'$ and wins the game if $b=b'$. We discuss how $A$ produces this output in Algorithm~\ref{alg:embedding_succ}\label{line:computebit}.
    \end{enumerate}
\end{framed}


\begin{algorithm}
   \caption{Training a sparse linear attack using shadow model training.}
   \label{alg:embedding_prep}
\begin{algorithmic}
   \STATE {\bfseries Input:} A target dataset $D^*$
   \STATE {\bfseries Output:} A linear attack model $l$ and a query set $W$
   \begin{enumerate}
   \STATE Following the notation of the security experiment in Figure~\ref{fig:userdsattack}, let the target tenant data $D^*$ be a sequence of words $D^*= \{ w_0, \ldots, w_s\}$.
   
   \STATE Sample multiple datasets $T_1, \ldots, T_k$ such that $D^* \in T_i$ from $\mathcal{D}_{\mathsf{shadow}}$
     \STATE Sample multiple datasets $T'_1, \ldots, T'_k$ such that $D^* \notin T'_i$ $\mathcal{D}_{\mathsf{shadow}}$
     \STATE Train multiple embeddings $m_1,\ldots, m_k, m'_1, \ldots, m'_k$ using the respective datasets
     \STATE Query the embeddings on all the words in ${D^*}$
     \STATE Compute the distances between embeddings of adjacent words $(w_0,w_1), (w_1,w_2), \ldots, (w_{s-1},w_s)$ for all embeddings $m_i$ and $m'_i$ to get $\sigma(m_i)=(||m_i(w_1)-m_i(w_0)||_2,\dots, ||m_i(w_s) - m_i(w_{s-1})||_2)$. \label{step:adjpairs}
     \STATE Find a linear function of these distances $l$ such that $l(\sigma(m_i))>0 \wedge l(\sigma(m'_i))<0$ from the following family of functions:
     $L = \left\{ l(x) = \beta + \sum_{j=1}^s \alpha_j x_j, \forall j \in \{1, \ldots, s\} \ \alpha_j \in  \mathbb{R}; \beta \in \mathbb{R}  \right\}$  
     by solving the following LASSO optimization problem:
     $$\min_{l \in L_{D^*}} \sum_{i=1}^k \Big(1-l\big(\sigma(m_i)\big)\Big)^2 + \Big(1+l\big(\sigma(m'_i)\big)\Big)^2 + \lambda |l|_1.$$ with appropriate parameter $\lambda.$
   \STATE Let $l=(\beta,\alpha_1,\dots,\alpha_{s})$. Then construct the query set $W=\{\text{all } w_i \text{ such that } \alpha_i\neq0 \text{ or } \alpha_{i+1}\neq 0 \}$.     
   \RETURN $l$ as the attack model and $W$ as query set.
   \end{enumerate}
\end{algorithmic}
\end{algorithm}

\begin{algorithm}
   \caption{Membership inference on target word embedding model.}
   \label{alg:embedding_succ}
\begin{algorithmic}
   \STATE {\bfseries Input:} A linear model $l$ and a query set $W$
   \STATE {\bfseries Output:} A prediction $b'$
   \begin{enumerate}
        \STATE Query $M$ on all points in $W$ and use that to calculate $l(\sigma(M))$. Note that the adversary only needs to query words in $W$ to calculate this quantity.
        {\IF{$l(\sigma(M))<0$}
        \RETURN $b'=0$
        \ELSE
        \RETURN $b'=1$
        \ENDIF}
   \end{enumerate}
\end{algorithmic}
\end{algorithm}

 
\myparagraph{What is the intuition behind this attack?} 
Our attacks have two phases: the first phase is the \emph{preparation phase} where the adversary prepares some queries for the target model (word embedding in this case) without any interaction with the target model. The second phase is the \emph{query phase} which is interactive. In this phase, the adversary sends queries to the target models, collects the responses and processes them. In the first phase, the attack tries to find correlation between words that are special to the target data. For instance, imagine there is an embedding model that uses the text in this paper for training an embedding model. It is natural to expect that when that happens, the distance between words "membership" and "inference" decrease compared to when this paper is not present (unless the most of the text in the dataset is about membership inference attacks.). Our attack tries to identify such pairs of words whose distance will significantly change upon including the target dataset. Note that in training the attack model there is a parameter $\lambda$ that we can control to change the number of important pairs of words found in the users email data. Because of unique properties of LASSO regression, we know that the final linear model is going to be sparse. This means that in the query phase the attacker will only need to query a few words and not all the words in the target dataset. By controlling the $\lambda$ parameter we can change the number of pairs of words that have non-zero weight in the linear combination. Changing $\lambda$ can also change the generalization of attack to unseen models. We describe the concrete details of the our attack in Section~\ref{ssec:embedding_attack}. 
 
\subsection{Attack results} \label{ssec:embedding_attack}

\begin{figure}
\begin{center}
    \includegraphics[scale=0.6]{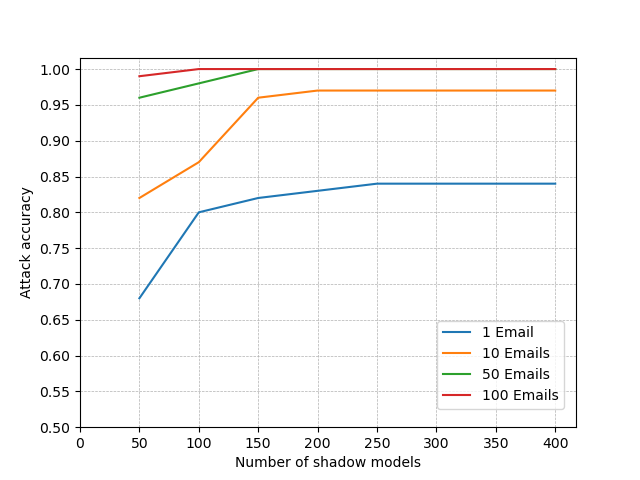}
    \caption{As numbers of shadow models increase, the attack becomes more successful in performing membership inference. In all setting, the attack seem to reach a plateau after 200 shadow models. Interestingly, the attack can success even when there is a single email in the target email set.}
    \label{fig:exp1}
\end{center}    
\end{figure}

    We use our attack against the Word2Vec algorithm \citep{rong2014word2vec} according to security experiment \ref{ssec:embedding_expt}. Figure \ref{fig:exp1} shows the success of the attack. For this experiment, we use Enron email distribution as the shadow distribution $\mathcal{D}_{\mathsf{shadow}}$, target dataset distribution and, the distribution of rest of the training data. In Figure \ref{fig:exp1}, we vary the size of target dataset. As expected, the attack works better with larger target datasets. Below, some details about this experiment is presented.
    
    \myparagraph{Experimental details} We use a subset of 10000 emails from Enron to train word embeddings using the Word2Vec algorithm. We also have a target email set with size (denoted by $s$) varying from 1 email to 100 emails. This means the distribution of the target email set is equal to the distribution of all other emails (In particular, in the security experiment \ref{ssec:embedding_expt}  $D^*$ is sampled from the same distribution $\mathcal{D}$).  The job of adversary is to guess whether the target email set has been used in the training set of the embedding model. Another variable is the number of (shadow) embedding models (denoted by $k$) that the attacker trains to optimize the attack model. 
    
    \myparagraph{Data Preprocessing:} We first preprocess the Enron and Avocado datasets by replacing all the words that are not in a public dictionary of words with a specific dummy word. For this purpose, we use the dictionary of a pretrained word embedding model on the Google News Corpus\footnote{https://github.com/mmihaltz/word2vec-GoogleNews-vectors}. Then both datasets are divided into two equal parts. One part for shadow model training that adversary has access to and another part for training the target embedding model.
    \myparagraph{Embedding parameters} We use Word2Vec \citep{rong2014word2vec} with window size 5. We use dimension 80 for the vector representation and iterate 20 times over all the data to train the Word2Vec models. The vocabulary of the model is set to the vocabulary of the entire prepossessed Enron dataset. Then, in the process of training, the algorithm is set to ignore all words that happen with frequency less than 20 in the training data.
    \myparagraph{Lasso Parameters:} Our attack uses a Lasso parameter of $\lambda=1/\sqrt{k}$ to ensure that the number of selected word pairs does not exceed 50. 
    
    \myparagraph{Randomness and Repetitions} In the upcoming figures, we repeat each experiment 50 times. We first sample 10 different target email sets from Enron distribution and for each target email set, we repeat the experiments security experiment 5 times. At the end, we report the average of the success of adversary in all experiments.
    
\subsection{Other variants of the attack}
The attack described above has three major assumptions that might be unrealistic in some scenarios. First, our attack is assumed to have sample access to the distribution of text data used for training the embedding.  Second, we assume that the target data that adversary aims at doing membership inference on is either completely used in the training set or not used at all. And third, we assume black box access to the embedding model. In the rest of this section, we discuss how our attack can still succeed in scenarios where the the first two assumptions are violated. Then, in Sections \ref{sec:classifier_attack} and \ref{sec:generative_security_exp} we discuss how the attack can still work even if it does not have access black-box access to an embedding.

\begin{figure}
\begin{center}
    \includegraphics[scale=0.6]{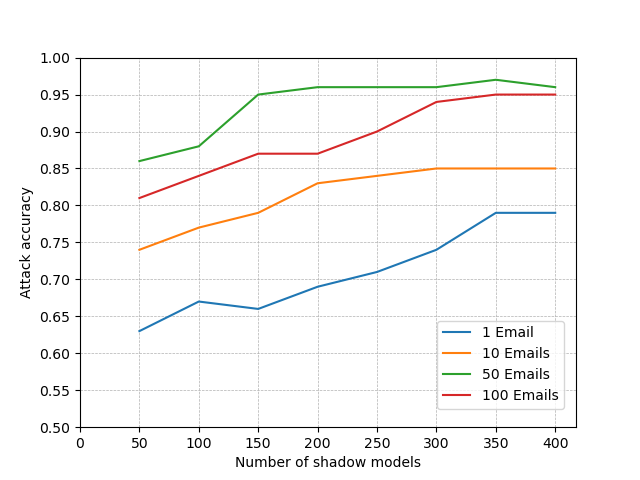}
    \caption{Success of the attack when the shadow models are trained on a proxy distribution (Avocado) instead of the original (Enron). Interestingly, the knowledge of the exact distribution is not crucial to the attack if enough number of shadow models are trained.}
    \label{fig:exp2}
\end{center}    
\end{figure}

\begin{figure}
\begin{center}
    \hspace{-10pt}\includegraphics[scale=0.6]{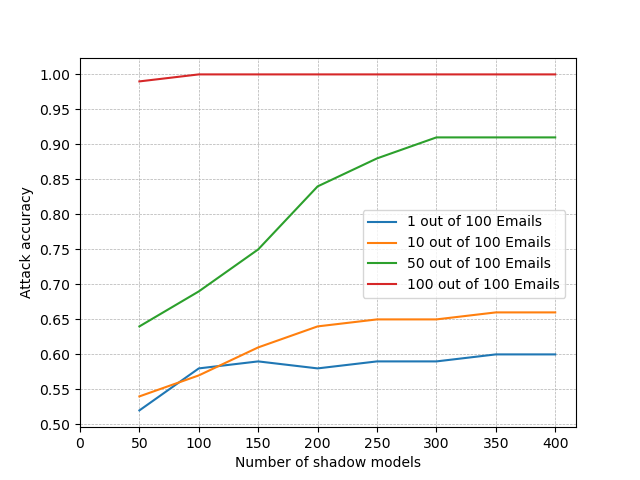}
    \caption{Success of attack when a fraction of target dataset is used in training. }
    \label{fig:exp3}
\end{center}    
\end{figure}

\myparagraph{Knowledge of the training distribution Assumption} In order to train shadow embeddings, it seems crucial for the adversary to have sampling access to the same distribution that the model trainer uses. Although this assumption is relevant in many scenarios, it will still be interesting to see how removing this assumption can affect the adversary. In order to understand the effect of this assumption on the attack, we perform experiments where the adversary does not have access to the same text distribution. In this set of experiments, the adversary uses the Avocado dataset to train shadow embeddings while the actual target model is trained on Enron dataset. Figure \ref{fig:exp2} shows the success of the attack in these experiments when enough number of shadow models are trained. All the parameters in this experiment is similar to the experiments done in Figure~\ref{fig:exp1}. 
\myparagraph{All-or-nothing Assumption} Another assumption behind the attack that might be violated is that the target data $D^*$ will be either completely used or will not be used at all. In real scenarios, the model trainer can sub-sample some data from each user and then train on that. This can potentially hurt the attack as the identified pair of words might not appear in the training set of the model at all. In order to account for this, we run an adaptive attack against this sub-sampling defense. In this attack, the adversary incorporates the sub-sampling step in the training algorithm used for training the shadow models. When this adaptive attack is used, the attack is has to select a more diverse set of word pairs so that if some of them did not appear in the actual training set, then other pairs can still help to perform the inference. Figure \ref{fig:exp3} shows the success of our attack in this setting. 
 \section{Embedding Attack on a Text-Classification Model} \label{sec:classification_attack}
\label{sec:classifier_attack}
Our embedding attack in Section \ref{sec:embed_attack} works when the adversary has black-box access to  the embedding. However, the embeddings might not be directly accessible by adversary. For example, adversary might only have access to a classifier that is trained on a public dataset which uses the embedding in its training process. In this section, we try to address this kind of setting. 

\subsection{Attack as per the Security Experiment}
The security experiment here is identical to the experiment in Section~\ref{ssec:embedding_expt} except that additional to $D^*$ it has another parameter $\mathcal{D}_c$ which a labeled data distribution that is used for training the classifier. Additionally the following steps are different:
\begin{itemize}
    \item $D^*$ is the email data set of a user of Avocado dataset.
    \item In Step~\ref{line:whichdist}, the $\mathcal{D}_i$ distributions are distribution of emails from Avocado dataset (instead of Enron).
    \item In Step~\ref{line:whichmodel}, after training Word2vec model $M$, a dataset $D_c\gets \mathcal{D}_c$ is sampled and then mapped to embedding space using $M$ to get $M(D_c)$. Then a spam classifier is trained on $M(D_c)$ to get a linear classifier $C$.
    \item In Step~\ref{line:prep}, the attacker adaptively queries the classifier $C$ (instead of the Embedding model) and prepares a set of query documents (instead of query words) using Algorithm~\ref{alg:classification_prep}.
    \item In Step~\ref{line:computebit}, the attacker runs Algorithm~\ref{alg:computebit} to decide on the bit $b'$.
\end{itemize}

We now propose our attack for this setting. 

\begin{algorithm}
   \caption{Training the attack using shadow model training.}
   \label{alg:classification_prep}
\begin{algorithmic}
   \STATE {\bfseries Input:} A target data $D^*$
   \STATE {\bfseries Output:} An attack model $l$ and a query set $S$.
   \begin{enumerate}
        \STATE perform the preparation step of the word embedding attack as described in Algorithm \ref{alg:embedding_prep} on target dataset $D^*$ to get the set of words $W$. Also keep the embeddings $m_1,\dots,m_k,m'_1,\dots,m'_k$.
        \item Sample $2k$ different datasets $T^c_1,\dots, T^c_{2k}$ from the classification distribution $\mathcal{D}_c$ and use embeddings to map them into the embedding space to get $m_1(T^c_1),\dots, m_k(T^c_k),m'_{1}(T^c_{k+1}),\dots,m'_{k}(T^c_{2k})$ . 
        \STATE Train $2\cdot k$ different linear classifiers $\{c_1,\dots,c_k, c'_{1},\dots, c'_{k}\}$ on the embedded datasets of previous step.
        
        \STATE create a set of queries by first sampling a set of random query documents $q_1,\dots,q_100$ from $\mathcal{D}_c$.
        \STATE For $q_i$, each word $w$ in the set of word pairs $W$ and each $0\leq j\leq 5$, create a query point $q_{(w,i,j)}$ which is same as $q_i$ with the difference that it is appended with $i$ repetitions of word $w$. let $S$ be the union of all $q_{w,i,j}$.
        \STATE Query each $c_i$ and $c'_i$ on $S$ to get a set of labels $L_i$ and $L'_i$ respectively.
        \STATE Create a dataset $\{(T_1,
        1),\dots,(T_k, 1), (T'_1,0), (T'_k,0) )\}$ and train an attack classifier $l$ on this dataset using a Random forest classifier.
        \RETURN $l$ as the attack model and $S$ as the set of queries.
\end{enumerate}

\end{algorithmic}
\end{algorithm}
\begin{algorithm}
   \caption{Membership inference on the target classification model.}
   \label{alg:computebit}
\begin{algorithmic}
   \STATE {\bfseries Input:} Attack model $l$ and a query set $S$
   \STATE {\bfseries Output:} A prediction $b'$
   \begin{enumerate}
   \STATE Query $C$ on $S$ to get a set of labels $L$.
   \RETURN $l(L)$.
\end{enumerate}

\end{algorithmic}
\end{algorithm}
 \myparagraph{Intuition behind the attack:} The main goal of the design of our classification attack is to extract the information from the embedding attack described in Section \ref{sec:embed_attack}. In particular, as we keep appending the special word $w_i$ to a document $d$, we might see some change in the prediction of classifier when the number of repetitions of the word exceeds some threshold $\sigma_i$. This threshold gives us a mean to measure the closeness of two words $w_i$ and $w_j$ by comparing the value of $\sigma_i$ and $\sigma_j$. Specifically, if word $w_i$ and $w_j$ are close together in the embedding space, we expect them to have similar behavior in changing the prediction of the classifier. Although this measure is different from the distance of the words in the embedding space, it still has enough information to make the attack succeed with proper shadow model training.
\subsection{Attack Success}

\myparagraph{Experiment details:} We evaluate our attack in a setting where the dataset used for embedding is different from dataset used for classification. In particular, the embedding data is sampled from Avocado dataset whereas the classification data is sampled from Enron dataset. We use 30 random users from Avocado dataset to train a word embedding. Then use that embedding to transfer emails in the Enron dataset into the embedding space. Then we train a linear classifier on embedded Enron dataset. On average, we get around $93\%$ accuracy for this classifier (See Table \ref{tab:classification_acc}). Now the goal of adversary is to guess if the training set for the underlying embedding include the target dataset. 
\esha{I thought we had the other setting too where the embeddings are the same? Do we still have that number? Also, Shall we refer back to $D_c$?}
\myparagraph{Experiments} Our attack uses 500 shadow embedding and 500 shadow classifiers and uses lasso parameter $0.1$ to come up with the set of words. Then after getting the set of words our adversary constructs a set of query points as described in the attack algorithm. Our final attack model is a random forest classifier applied on the shadow classifiers. Table \ref{tab:classification} shows the success of our attack.

\begin{table}
\vskip 0.15in
\begin{center}
\begin{sc}
\begin{tabular}[t]{p{0.3\textwidth}p{0.3\textwidth}}
\toprule
target dataset size & Attack Accuracy \\
\midrule
10 & 0.65{\footnotesize{$\pm$8.15}} \\
100 & 0.69{\footnotesize{$\pm$5.08}} \\
1000 & 0.79{\footnotesize{$\pm$3.75}} \\
\bottomrule
\end{tabular}
\end{sc}
\caption{Attack success v.s. number of sentences in the target dataset. In these experiments, Avocado data is used for training the embedding and the Enron data is used for training the classifier.}\label{tab:classification}
\end{center}

\end{table}


We also try another variant of the attack where the data used for classification and embedding is the same. This setting will be relevant in scenarios where the word embedding and classifier are trained together. The security experiment for this setting will be similar to the classification experiment except that there is no dataset for the classification algorithm. Instead, the classifier is trained on the same data that is used for training the embedding. Similarly, in the algorithm for training the shadow models, the adversary will use the same sampled dataset to train the shadow models. This variant of the attack has less entropy and the attack is expected to be more successful. Table \ref{tab:classification_2} shows the success of attack in this setting.

\begin{table}
\vskip 0.15in
\begin{center}
\begin{sc}
\begin{tabular}[t]{p{0.3\textwidth}p{0.3\textwidth}}
\toprule
target dataset size & Attack Accuracy \\
\midrule
10 & 0.77{\footnotesize{$\pm$5.12}} \\
100 & 0.84{\footnotesize{$\pm$2.35}} \\
1000 & 0.91{\footnotesize{$\pm$1.78}} \\
\bottomrule
\end{tabular}
\end{sc}
\caption{Attack success v.s. number of sentences in the target dataset. These experiments capture the setting that the same data is used to train the classification and embedding models.}\label{tab:classification_2}
\end{center}

\end{table}

\begin{table}
\vskip 0.15in
\begin{center}
\begin{sc}
\begin{tabular}[t]{p{0.08\textwidth}p{0.2\textwidth}p{0.18\textwidth}p{0.18\textwidth}}
\toprule
Model & Embedding dataset  & classification accuracy (train) & classification accuracy (test)\\
\midrule
SVM & Enron & 95.18{\footnotesize{$\pm$2.12}} & 94.27{\footnotesize{$\pm$1.05}}\\
SVM & Avocado & 93.98{\footnotesize{$\pm$1.95}} & 93.34{\footnotesize{$\pm$1.08}}\\

\bottomrule
\end{tabular}
\end{sc}
\caption{Accuracy of the target classifier.}\label{tab:classification_acc}
\end{center}

\end{table}
 \section{Embedding Attack on a Text-Generation Model}\label{sec:tg_attack}

In the previous section, we show an application of our embedding attack to a text-classification model. In this section, we extend our approach to attack a text-generation model. We focus on the next-word prediction task as the text-generation setting. This is a very popular setting in which language models have been deployed in practice to perform text auto-completion in emails and predictive keyboards \citep{swiftkey, gmailsc}. On the other hand, such models are extensively trained on personal data, e.g. users' emails, documents, chats etc., which may lead to privacy leakages as studies show in \cite{song19, secretsharer, carlini2020extracting, inan2021}.

\subsection{Attack as per the Security Experiment}
\label{sec:generative_security_exp}

\myparagraph{Dataset} We use the Avocado dataset~\citep{avocado2015} in our experiments. The dataset consists of 279 users. We split up the dataset in two parts: $\Dtrain$ and $\Dshadow$ where each part contains 100 users' data picked randomly from the dataset. $\Dshadow$ is only used in the preparation phase of the attack. $\Dtrain$ is further divided in two parts as $\Din$ and $\Dout$ randomly, each containing data of 50 users. The target model trains on $\Din$ and the attack experiment is performed on both $\Din$ and $\Dout$.

\myparagraph{Target model}
Similar to the setting in \cite{song19}, we use long short-term memory (LSTM) \citep{lstm} in our language model. LSTM is a special type of Recurrent Neural Networks (RNNs) that can capture the long-term dependency in the text sequence. In this network, the text sequence of tokens is first mapped to a sequence of embeddings. The embedding is then fed to the LSTM that learns a hidden representation for the context for predicting the next word. We use a two-layer LSTM model as the language model for the next-word prediction task. We set both the embedding dimension and LSTM hidden-representation size to 500 (gives around 50 million parameters). All model weights including the embeddings are initialized randomly before training. We use the Adam optimizer with the learning rate set to 1e-3 and batch size to 64. We repeat our experiments over five random runs. The performances of the models are presented in Table \ref{table:lm_perf}.

\begin{table}
\vskip 0.15in
\begin{center}
\begin{sc}
\begin{tabular}[t]{p{0.18\textwidth}p{0.15\textwidth}p{0.15\textwidth}}
\toprule
Model & Train ppl & Test ppl \\
\midrule
2-layer LSTM & 94.70{\footnotesize{$\pm$3.16}} & 107.74{\footnotesize$\pm$2.55} \\
\bottomrule
\end{tabular}
\end{sc}
\end{center}
\caption{Performances of the target models. The mean and standard deviation are given over five runs. Ppl stands for perplexity. }
\label{table:lm_perf}
\vskip -0.1in
\end{table}

Recent work has shown both successful extraction of training data \citep{snapshotattack, carlini2020extracting, inan2021} and membership inference \citep{song19} in language models. The membership inference attack we describe in this section is different in the sense that our method only uses the top-1 predictions of the model, i.e. our attack is label-only, which is a strong and realistic attack setting that has been explored recently in visual domain as well \citep{choo2020labelonly, li2021membership}. Another advantage of our attack is that it does not require any shadow training as opposed to \cite{song19}, hence, the target model need not be assumed to be known and the attack is computationally efficient. We next describe the attack setting as per the security experiment.
 
 \begin{framed}
    \begin{enumerate}
        \item  In this experiment, each of the $n=100$ distributions $\mathcal{D}_i$ for $i = 1, \ldots n$ is a user's email distribution from the Avocado dataset.
        \item $D_i \gets \mathcal{D}_i$.  Here $D_i$ is user $i$'s data.
        \item Run the following several times:
        \begin{enumerate}
            \item Initialize a bit vector $\textbf{b}= b_1, \ldots, b_n = 0^n$
            \item The dataset $\Dtrain$ is split in two parts: $\Din$ and $\Dout$ where each part contains data of 50 users. $\Din$ and $\Dout$ are constructed as follows: The user id's in $\Dtrain$ are randomly permuted and then the first 50 users constitute $\Din$ and rest constitute $\Dout$. Let us denote the datasets in $\Din$ as $D'_1, D'_2,\ldots, D'_{50}$.
            \item Set the corresponding bits in $\textbf{b}$ to 1. 
            \item Train a model $M\gets L(D'_1, D'_2,\ldots, D'_{50}) $.
            \item For $i\in [n]$:  Run the attack as follows:
            \begin{itemize}
                \item The attacker is given $D_i$ for $i \in [n]$.
                \item The attacker has also $\aux = \Dshadow$
                \item $A$ adaptively queries the model $M$ on a sequence of points. The points are chosen as follows:
                    \begin{enumerate}
                    \item Preparation: This phase is identical to the preparation phase described in Section~\ref{sec:embed_attack}. The shadow data used is $\Dshadow$. Let $\mathcal{W_{\mathsf{u}}}$ be the words output by the preparation phase for user $\mathsf{u} \in \Dtrain$. 
                    \item For each user $\mathsf{u} \in \Dtrain$: Apply Algorithm \ref{alg:mia_wout_threshold} with $\mathcal{W_{\mathsf{u}}}$ to get $b'_\mathsf{u}$. \label{line:5computebit}
                \end{enumerate}
                \item The attacker outputs $b'_i$ indicating whether it thinks $D_i$ was included in the training set.
            \end{itemize}
            \item  Compute S=  $\sum_{i=1}^{n} (b'_i = b_i)/100$
        \end{enumerate}
        \item We compute a metric for the success of the attack by averaging the values of $S$ obtained in each of the runs.  
    \end{enumerate}   \label{LM:security_exp}
\end{framed}

Let us expand the Step~\ref{line:5computebit} of the security experiment introduced above. For each user in the target dataset, we apply the attack described in Section \ref{sec:embed_attack} using the shadow dataset, which generates a list of word pairs $(w_i, w_{i+1})$. Let us denote this list as $\mathcal{W}$. The attack operates on each pair $(w_i, w_{i+1})$ in a simple way by using the sequence that has this pair $(w_i, w_{i+1})$, querying the model with the context of up to and including $w_i$ and checking if the top prediction returned by the model is $w_{i+1}$. If there are multiple sequences that has the pair $(w_i, w_{i+1})$, the operation is performed on all of them. If there exists a pair of words in $\mathcal{W}$ satisfying this condition, the corresponding user is predicted as member. Otherwise, the user is predicted as non-member. The attack is described in Algorithm \ref{alg:mia_wout_threshold}.

\begin{algorithm}
   \caption{Label-only membership inference attack performed in Step~\ref{line:5computebit} of the security experiment.}
   \label{alg:mia_wout_threshold}
\begin{algorithmic}
   \STATE {\bfseries Input:} A language model $LM(\cdot)$ and the attack data $\mathcal{W_{\mathsf{u}}}$ of user $\mathsf{u} \in \Dtrain$
   \STATE {\bfseries Output:} The membership prediction $b'_\mathsf{u}$
   \FOR{$(w_i, w_{i+1})$ {\bfseries in} $\mathcal{W_{\mathsf{u}}}$}
   \STATE Construct $L = \{s \in {D}_u : (w_i, w_{i+1}) \in s\}$ where $s$ are sentences that contain $(w_i, w_{i+1})$
   \FOR{$s$ {\bfseries in} $L$}
   \STATE Find the index $j$ such that $s_j = w_i$
   \STATE Obtain the next-word prediction $p = LM(s_1, \ldots, s_j)$
   \IF{$p = w_{i+1}$}
   \RETURN $b'_\mathsf{u} \gets 1$
   \ENDIF
   \ENDFOR
   \ENDFOR
   \RETURN $b'_\mathsf{u} \gets 0$
\end{algorithmic}
\end{algorithm}

\subsection{Attack results} 
\label{sec:mia_wout_threshold}

We note that for a member prediction, it is sufficient to have at least one pair of words satisfying the described condition. Therefore, a well-curated list of word pairs should be generated so that the condition is not satisfied for any of the word pairs for a non-member and it is satisfied for at least one pair for a member. The approach described in Section \ref{sec:embed_attack} is effective in the sense that it can generate such well-crafted list of word pairs for each user. The results of the security experiment, which has been repeated five times with different randomness, are presented in Table \ref{table:mia_wout_threshold}.

We observe that with label-only access and not training any shadow models, our attack performs impressively. This is the first attack in such a setting to the best of our knowledge. For comparison, the attack in \cite{song19} under label-only access boils down to calculating the average accuracy over each user's data and selecting a threshold to decide membership using shadow model training. Without the latter step, the attack would naturally result in all-member prediction since the attack essentially uses all word pairs in a user’s data and each user has at least one word pair satisfying the required condition. This is equivalent to the first baseline in Table \ref{table:mia_wout_threshold} where we apply the attack in Algorithm \ref{alg:mia_wout_threshold} using all word pairs in a user's dataset. We point out another advantage of our method that it is query-efficient in the sense that it uses a curated list of word pairs, therefore, this substantially reduces the number of queries to the model compared to \cite{song19}. This means that our attack cannot trivially be mitigated by limiting the number of queries made to the model.

As a second baseline, we try to obtain a curated list of word pairs by excluding common word pairs if they are present in a public dictionary (we use GloVe public dictionary \citep{pennington2014glove}). As noted in \cite{song19} successful prediction of rare words provide strong signal for membership inference. Although this improved the attack performance as expected, it still substantially falls short of our attack. The results in Table \ref{table:mia_wout_threshold} show that our approach indeed generates a well-curated list of word pairs that provides good signal for membership inference. We finally highlight the fact that our attack, which is based on Word2Vec embedding, transfers through the LSTM-based language model, which trains its own embeddings different from Word2Vec. Therefore, we expect our MI attack to generalize to other text-generation models such as ones that are based on large transformer \citep{transformer} models for which shadow model training would be highly costly. 

\begin{table}
\vskip 0.15in
\begin{center}
\begin{sc}
\begin{tabular}[t]{p{0.15\textwidth}p{0.15\textwidth}p{0.15\textwidth}p{0.15\textwidth}}
\toprule
Attack & Accuracy & Precision & Recall \\
\midrule
Our attack & 0.708{\footnotesize{$\pm$0.023}} & 0.67{\footnotesize{$\pm$0.025}} & 0.82{\footnotesize{$\pm$0.047}} \vspace{.2cm} \\ 
Baseline 1 & 0.5{\footnotesize{$\pm$0.0}} & 0.5{\footnotesize{$\pm$0.0}} & 1.0{\footnotesize{$\pm$0.0}} \vspace{.2cm} \\ 
Baseline 2 & 0.637{\footnotesize{$\pm$0.032}} & 0.69{\footnotesize{$\pm$0.042}} &  0.478{\footnotesize{$\pm$0.066}} \\ 
\bottomrule
\end{tabular}
\end{sc}
\end{center}
\caption{Attack results of Section \ref{sec:generative_security_exp} over five runs (mean \& standard deviation). Baseline 1 is applying Algorithm \ref{alg:mia_wout_threshold} for {all word pairs} in a user's data. Baseline 2 is applying Algorithm \ref{alg:mia_wout_threshold} for all word pairs except the ones that can be found in a public dictionary (GloVe).}
\label{table:mia_wout_threshold}
\vskip -0.1in
\end{table}

 \section{Related work}\label{sec:related}

Embeddings, as a representation of the string format of text, has been one of the key stepping stones to successful machine learning models in NLP applications. Utilizing unsupervised learning on large corpus of text, various methods such as Word2Vec \citep{mikolov2013efficient}, Glove \citep{glove}, fastText \citep{fasttext} have been designed to produce a vector space where each word in the corpus is assigned a vector (embedding) in the generated space. These word embeddings can be more compact and maintain semantic similarity by being in close proximity to one other in the vector space, therefore, more favorable than one hot encoded vectors. Embeddings are the first block of the deep neural models such as ones that are based on recurrent neural networks (RNNs) \citep{mikolov10,Sundermeyer_lstmneural} or based on self-attention mechanisms of the transformer \citep{transformer}. These models have been widely employed in NLP applications and the granularity of the embedding can be word level or sub-word level (e.g. BPE tokenization \citep{bpe}) depending on the model architecture. 

When machine learning models are trained on sensitive personal data, utility as the performance of the model should not be the only metric of attention. In fact, a wide body of work has demonstrated privacy issues for machine learning models trained on personal data. In general, it is known that deep learning models can achieve perfect accuracy even on randomly labeled data \citep{zhang2016}. Strong memorization ability may actually be required to achieve near-optimal accuracy on test data when the data distribution is long-tailed as recently shown by \citet{feldman2020,brown2020memorization}. There are serious implications of this in the NLP domain that may lead to privacy breaches. For instance, \citet{carlini2020extracting} demonstrated that individual training examples from the GPT-2 language model \citep{radford2019language} can be recovered verbatim. In a transfer learning setup, \citet{snapshotattack} have shown that by having simultaneous black box access to the pre-trained and fine-tuned language models, rare sequences from the typically more sensitive fine-tuning dataset can be extracted successfully. In this regard, recent work \citep{secretsharer, inan2021, Fatemehsadat21} have also proposed metrics and mitigations to evaluate generative models from the perspective of privacy.

In case of classification tasks, label-only access may obstruct such a direct leakage from the training data. However, an indirect leakage of what is called membership inference attack \citep{shokri17} can still lead to privacy violations \citep{murakonda2020ml}. In a membership inference attack, the goal is to determine if a particular data point or a targeted user belongs to the training set of the model. There has been substantial progress in this area over a wide range of applications under different assumptions/settings \citep{shokri17, yeom18,  song19, nasr19, long18, LOGAN, truex18, irolla19, hisamoto20, salem18, sablayrolles19a, Leino20, choo2020labelonly}. 

While the settings in the aforementioned work are different than what we are focusing on in this work, we highlight the article~\cite{embeddingccs} as it is closest to this work. \cite{embeddingccs} studies information leakage from embeddings and we compare it with our work in great detail in the next section.

\subsection{Comparison with~\cite{embeddingccs}} \label{sec:compare}

In~\cite{embeddingccs}, the authors consider three types of attacks: 1) embedding inversion 2) sensitive attribute inference and 3) membership inference (MI). In this paper, we focus on MI and how it can be further used in downstream tasks. Like in the previous paper, we also focus on text input data. We will compare our attacks with the MI attacks in~\cite{embeddingccs}.

In~\cite{embeddingccs}, the authors address the following question: \emph{can an adversary with access to the embeddings extract the encoded sensitive information?}  The MI attack on word embeddings that the authors propose also has several differences from ours.

We go beyond what \cite{embeddingccs} consider as sensitive information: we also consider the question of what kind of sensitive information  an adversary can learn given black-box access to a model that uses embeddings as its first layer (rather than direct access to the embedding itself). 

\myparagraph{Definition of MI} The paper points out that unlike supervised learning, word embeddings trivially allows for word level MI: every word in the vocabulary is trivially a member of the training dataset $D_{\mathsf{train}}$. So, to meaningfully talk about MI, the definition of MI needs to be expanded. The way the authors expand on this is the following: They consider that the adversary has a target string of words $[w_1,\ldots, w_n]$, referred to as a context, and access to the embedding. The adversary tries to decide the membership for the context string. 

This definition, while interesting, is a little limited, in that the attack is very tied to the length of the input. For example, it's not clear how it would apply to membership inference for inputs that may have different length. On the other hand our security notion could apply to membership of substrings of specific length, but it can also apply to membership of emails/documents, or of a users' email or document collections, without any assumption that all users' emails/collections must have the same number of words.  


\myparagraph{Assumption of adversary's knowledge} In the MI attack in~\cite{embeddingccs}, the attacker has access to some auxiliary data labeled with membership. Moreover, in their attack, they show that their adversary's advantage increases as the central word in the 5-window context becomes rarer. Therefore, the success of the attack requires the adversary to pick a context whose central word is rare in the training distribution, which implicitly assumes that the adversary has enough knowledge of the training distribution to find these rare words. Moreover, success of their attack depends on finding these specially constructed context strings.


This is in contrast to ours, where the adversary does not have have access to any auxiliary labeled data. In fact, we show that, the adversary does not even need sample access to the exact training distribution of the target model; access to a similar distribution can be sufficient. Relaxing the assumption on the adversary's knowledge makes our attack stronger. 

\myparagraph{Success metric} \cite{embeddingccs} show that their attack is successful for certain types of contexts (specifically those whose central word is rare).  This is interesting in that it says specifically which types of substrings are vulnerable, but it leaves open the question of how common such substrings are in real datasets.

Our analysis, in contrast, shows that our attack is successful for input data drawn from real datasets.

\myparagraph{Attack specification}
The attack defined in~\cite{embeddingccs} is threshold based: they measure average cosine similarity of all word pairs in the context and sees if it is over a certain threshold. However, in the experiment they do not discuss how they compute the threshold. Deciding on a threshold is non-trivial and probably requires some additional shadow model training.

In contrast to this, in our attacks that use thresholds, we specify exactly how the  thresholds would be computed, so this cost is already included in the analysis of our attack.  Our text-generative model attack has the advantage that it doesn't require a threshold.   

\myparagraph{Fundamental attack intuition} The attack in~\cite{embeddingccs} is fundamentally based on the insight that rarer words are memorized more. They check with sliding window of size 5 with varying frequency of the central word and show that adversary's advantage increases as the central word becomes rarer. Their sentence embedding MI uses a similar idea.

This is fundamentally different from the intuition of our attack. We exploit the following property of embedding: a good embedding is expected to capture semantic relationships between words. This is widely believed to be an inherent property of a good embedding function \citep{schnabel2015}.   We combine this with the assumption, inherent in the way that embeddings are trained, that words that are adjacent/close in the training data will be somehow semantically close.

 \section{Conclusion}\label{sec:conclusion}

In this paper we looked at a  simple embedding function that lies at the heart of almost any NLP application, namely, word embedding. First we show that word embeddings are vulnerable to black-box membership inference attack. Then we show that this leakage persists through two other major NLP applications: classification and text-generation, even when the embedding layer is not exposed to the attacker. Our attacks exploit a property of word embeddings (preserving semantic relationship of words), which is widely believed to be a property of a good embedding function. Whether this leakage is in some way inherent is a question that requires further investigation.

A secondary contribution of our attack is a cheaper membership inference on text-generative models, which does not require any expensive training of text-generative models as shadow models.

Given the vulnerability of word embeddings and its persistence through other downstream tasks that our work exposes, and the ubiquitous use of word embeddings in almost all NLP tasks, we believe our work raises an important question of how to mitigate such vulnerability. Applying differential privacy \citep{dwork2011differential} in training \citep{abadi2016deep} will protect the model as a whole, hence including the embedding layer. Therefore, this is the most obvious possible defense here, but differentially private model training is substantially slower in comparison and may affect the utility of the model negatively \citep{disparate}. While this line of defense is being actively researched, it is also worth investigating other heuristic defenses that could cater to specific applications.

We conclude the article with two possible lines of future work. The first one is to measure the effectiveness of heuristic defenses or differential privacy with realistic privacy budget in defending against our attack. The second is to apply our attack to other models such as those based on transformers.
\bibliographystyle{ACM-Reference-Format}
\bibliography{references}

\end{document}